\documentclass[11pt]{article}

\usepackage[preprint]{acl}

\usepackage{times}
\usepackage{latexsym}
\usepackage{makecell}
\usepackage{listings}
\usepackage{booktabs}
\usepackage{minted}
\usepackage[T1]{fontenc}

\usepackage[utf8]{inputenc}

\usepackage{microtype}
\usepackage{hyperref}
\usepackage{inconsolata}

\usepackage{graphicx}

\usepackage{tcolorbox}
\usepackage{xcolor}

\definecolor{terminalbg}{RGB}{28,28,28}
\definecolor{terminaltext}{RGB}{230,230,230}

\tcbset{
terminalstyle/.style={
    colback=gray!15,
    colframe=black,
    coltext=black,
    arc=4pt,
    boxrule=0.5pt,
    left=6pt,
    right=6pt,
    top=6pt,
    bottom=6pt
}
}
\title{DEP: A Decentralized Large Language Model Evaluation Protocol}
\author{
  {\bf Jianxiang Peng$^{1}$\thanks{Equal contribution}, Junhao Li$^{1}$\footnotemark[1], Hongxiang Wang$^{1}$\footnotemark[1], Haocheng Lyu$^{1}$\footnotemark[1], Hui Guo$^{1}$}\\
  {\bf Siyi Hao$^{1}$, Zhen Wang$^{1}$, Chuang Liu$^{2}$, Shaowei Zhang$^{1}$, Bojian Xiong$^{1}$, Yue Chen$^{1}$}\\
  {\bf Zhuowen Han$^{1}$, Ling Shi$^{1}$, Tianyu Dong$^{1}$, Juesi Xiao$^{1}$, Lei Yang$^{1}$}\\
  {\bf Yuqi Ren$^{1,3}$\footnotemark[2], Deyi Xiong$^{1,3}$\thanks{Corresponding authors}}\\
  $^{1}$TJUNLP Lab, Tianjin University, Tianjin, China\\
  $^{2}$National Supercomputing Center in Tianjin\\
  $^{3}$Yuanhui AI\\
  \texttt{\{pjasonx, ryq20, dyxiong\}@tju.edu.cn}\\
}

\begin{document}
\maketitle
\begin{abstract}
With the rapid development of Large Language Models (LLMs), a large number of benchmarks have been proposed. However, most benchmarks lack unified evaluation standard and require the manual implementation of custom scripts, making results hard to ensure consistency and reproducibility. Furthermore, mainstream evaluation frameworks are centralized, with datasets and answers, which increases the risk of benchmark leakage. To address these issues, we propose a \textbf{D}ecentralized \textbf{E}valuation \textbf{P}rotocol (\textbf{DEP}), a decentralized yet unified and standardized evaluation framework through a matching server without constraining benchmarks. The server can be mounted locally or deployed remotely, and once adapted, it can be reused over the long term. By decoupling users, LLMs, and benchmarks, DEP enables modular, plug-and-play evaluation: benchmark files and evaluation logic stay exclusively on the server side. In remote setting, users cannot access the ground truth, thereby achieving data isolation and leak-proof evaluation. To facilitate practical adoption, we develop \textbf{DEP Toolkit}, a protocol-compatible toolkit that supports features such as breakpoint resume, concurrent requests, and congestion control. We also provide detailed documentation for adapting new benchmarks to DEP. Using DEP toolkit, we evaluate multiple LLMs across benchmarks. Experimental results verify the effectiveness of DEP and show that it reduces the cost of deploying benchmark evaluations. As of February 2026, we have adapted over 60 benchmarks and continue to promote community co-construction to support unified evaluation across various tasks and domains.

\end{abstract}

\textbf{Website:} \url{http://dep.openeval.org.cn}


\section{Introduction}
As LLMs continue to advance, previous benchmarks quickly become saturated, driving the continuous emergence of new evaluation tasks and datasets. In ACL 2025 \citep{acl-ws-2025-long} and EMNLP 2025 \citep{emnlp-2025-main} alone, nearly 700 benchmark-related studies have emerged. However, most benchmarks are released independently, with heterogeneous data formats, inference pipelines, and evaluation scripts, lacking a unified evaluation interface specification \citep{ni2025surveylargelanguagemodel}. As a result, model developers who wish to evaluate on a given benchmark, usually need to implement custom inference scripts or manually adapt the benchmark to existing evaluation platforms. This not only increases time cost but also makes it difficult to ensure the consistency and reproducibility of results. 

Additionally, existing evaluation frameworks only cover a few popular benchmarks, offering limited support for newly released benchmarks or domain-specific evaluations \citep{liu-etal-2024-openeval,he-etal-2024-ultraeval}. They follow a centralized design that integrates inference logic, datasets, evaluation metrics, and execution control. Such frameworks are large-scale and high coupled, with limited pluggability and a high learning cost \citep{yao-etal-2025-value,li-etal-2023-cleva,he-etal-2024-ultraeval}. Benchmark builders are frequently required to restructure their data and logic according to platform-specific formats, which restricts the diversity of evaluation methods. Furthermore, centralized frameworks need to integrate ground truth for evaluation, leaving the risk of test set leakage and data contamination.

\begin{figure}[!t]
    \centering
    \includegraphics[width=\linewidth]{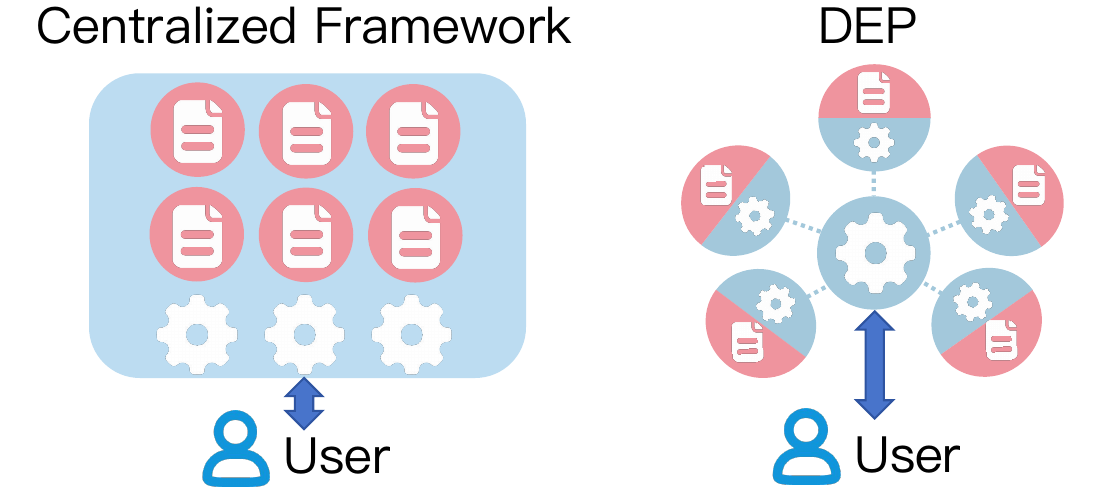}
    \caption{Comparison between centralized framework and our DEP. Red denotes benchmark files while blue evaluation modules.}
    \label{Figcentral}
\end{figure}

To address these challenges, we propose a \textbf{D}ecentralized \textbf{E}valuation \textbf{P}rotocol (\textbf{DEP}) that encourages LLM community co-construction. Figure \ref{Figcentral} illustrates the differences between DEP and the centralized framework. DEP does not enforce mandatory constraints on the original benchmark data format. Instead, by building a matching benchmark server script for each dataset, DEP implements a unified, standardized evaluation interface, enabling compatibility between different models, datasets, and evaluation methods. Architecturally, DEP decouples LLMs, clients, and benchmarks. In DEP, a client communicates with benchmark servers only through the protocol-defined standard interface, while evaluation logic and benchmark artifacts are contained entirely on the server side. If the benchmark authors intend to make the ground-truth answers public, they can release the corresponding scripts for local mounting, without any additional work. Correspondingly, these same servers can also be deployed remotely. Developers may either deploy them independently or host them on the benchmark server platform. We will provide a dedicated platform, where hosting can be requested through our website.

Furthermore, we implement \textbf{DEP Toolkit}, an evaluation tool that supports large-scale LLM benchmarking with breakpoint resume, concurrent requests, and congestion control. It works with both local and API-based inference, along with multiple evaluation methods and metrics, including custom-defined metrics and LLM-as-a-judge.

We conduct systematic evaluations of various open-source and closed-source LLMs using the DEP toolkit. Experimental results demonstrate the effectiveness of DEP while reducing the cost of benchmark integration and reuse.

In summary, our contributions are as follows: 

\begin{itemize}
    \item[$\bullet$]We introduce DEP, a decentralized evaluation protocol that decouples benchmarks from evaluation clients. Compared to centralized evaluation, DEP offers the advantages of zero-code convenience, reproducible evaluation, data privatization, and community-driven sharing.

    \item[$\bullet$]DEP allows benchmark authors to integrate their benchmarks without changing the original dataset format. The protocol is compatible with various evaluation set formats and inference methods, supporting a wide range of evaluation metrics and methodologies.

    \item[$\bullet$]We release DEP toolkit, a comprehensive toolkit implementing the protocol, with practical features for large-scale evaluation.

    \item[$\bullet$]We provide protocol and development documentation to facilitate the rapid adaptation. As of February 2026, we have adapted over 60 benchmarks to DEP.\footnote{http://dep.openeval.org.cn/News}
\end{itemize}

\begin{figure*}[!t]
    \centering
    \includegraphics[width=\linewidth]{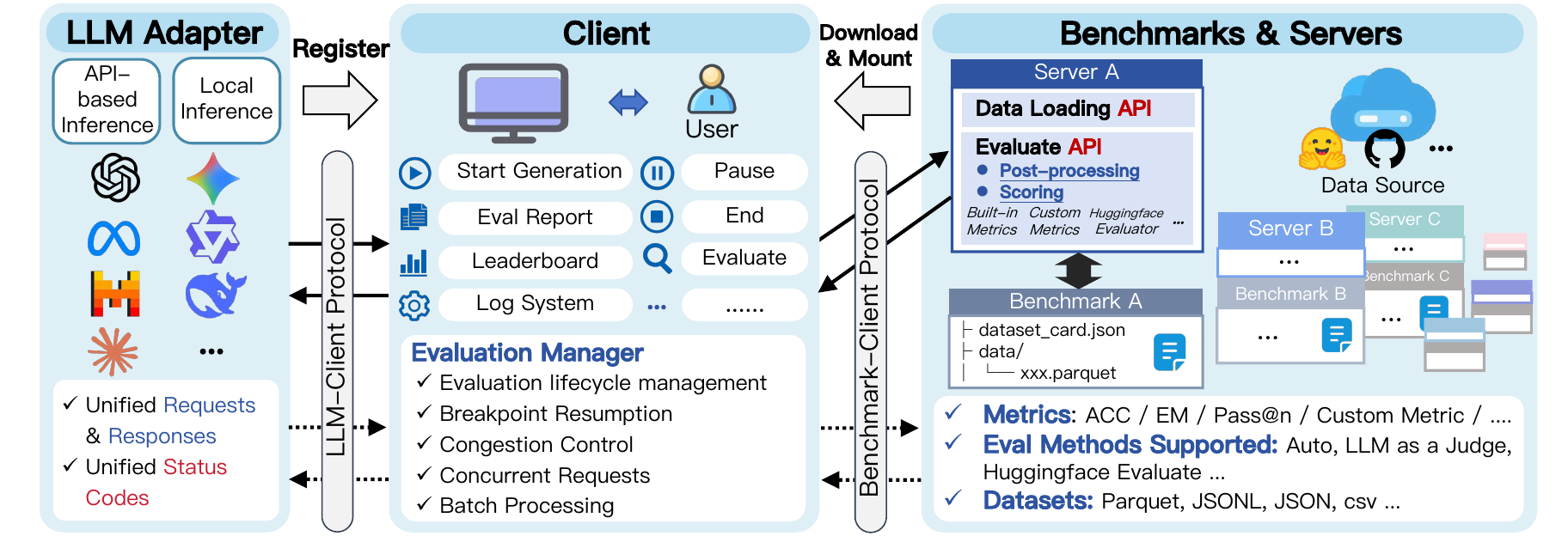}
    \caption{Overview of DEP. The client connects to the LLMs and benchmarks via a unified interface. The LLM Adapter standardizes various LLM interfaces and inference methods, while the benchmark server handles requests from the client, such as data loading and evaluation submission.}
    \label{Figure1}
\end{figure*}

\section{Related Work}
\paragraph{LLM Benchmarks.} With the continuous advancement of LLM capabilities, previous benchmarks are gradually becoming obsolete, making it difficult to effectively differentiate performance across LLMs. To characterize model performance in finer detail across domains such as knowledge \citep{hendrycks2020measuring,zhang2023evaluating}, code generation \citep{chen2021evaluating,zhuo2024bigcodebench}, and safety alignment \citep{mazeika2024harmbench,wang2023not}, researchers are constantly proposing new evaluation tasks and datasets \citep{ni2025surveylargelanguagemodel,shen2023large}. This has led to a highly fragmented evaluation ecosystem. Most benchmarks are released as independent repositories with disparate data formats, inference methods, and evaluation scripts, leaving the issue of standardized evaluation interfaces unresolved. DEP proposes a unified protocol, establishing a standardized interface without restricting the underlying expression or format of a benchmark.

\paragraph{Evaluation Frameworks.} The LLM community has introduced various centralized evaluation frameworks. For instance, LM Evaluation Harness\footnote{https://github.com/EleutherAI/lm-evaluation-harness} provides a unified task encapsulation and model adaptation interface, significantly lowering the barrier to entry for mainstream benchmarks. Similarly, OpenAI Evals\footnote{https://github.com/openai/evals} supports flexible evaluation definitions based on templates and scoring functions. Other examples include frameworks released by \citet{liu-etal-2024-openeval} and \citet{yu-etal-2024-freeeval}. Hugging Face Evaluate\footnote{https://github.com/huggingface/evaluate} integrates implementations for various common metrics. However, these frameworks typically adopt a centralized architecture, integrating datasets, LLM inference logic, and evaluation logic within a single platform. This results in bloated systems with poor scalability. Benchmarks often must be refactored to fit the platform’s specific data formats and interface specifications, which limits flexibility. For example, while LM Evaluation Harness supports new datasets via YAML files, users must adhere to strictly constrained fields. Meanwhile, OpenAI Evals does not support custom evaluation scripts. In contrast, DEP does not aim to build yet another evaluation platform. Instead, it introduces a protocol-level mechanism that allows benchmarks to be evaluated in a decentralized manner, as simply as API call.

\paragraph{Preventing Data Contamination.} As training corpora expand, test set contamination has drawn increasing scrutiny \citep{ni2025training}. If ground-truth answers are publicly distributed, LLMs can indirectly encounter these data during training phases, which undermines the credibility of the evaluation results \citep{balloccu-etal-2024-leak,fang-etal-2025-lastingbench}. To mitigate this risk, some organizations employ a private or online evaluation, requiring developers to submit LLM’s generation results, the evaluator then calculates scores and announces rankings in a closed environment \citep{wang-etal-2018-glue, huang2023c, liu-etal-2024-openeval}. While this approach reduces data contamination, it often relies on manual coordination or platform-specific implementations and lacks a unified technical protocol. DEP addresses this by supporting a ground truth isolation mechanism directly at the architectural level.

\section{DEP Design}

DEP employs a three-component architecture consisting of the LLM adapter, the client, and the benchmark server. As shown in Figure \ref{Figure1}, the LLM adapter provides a unified LLM calling interface to mask differences between various LLM inference methods and APIs. The client is responsible for managing the evaluation lifecycle, scheduling tasks, and controlling concurrency and rates. The benchmark server handles evaluation submissions from the client, implementing evaluation and metrics logics.

\subsection{LLM Adapter}
DEP defines a unified calling interface within the LLM adapter. This structure allows the client to complete scheduling and evaluation without perceiving specific LLM types.

\textbf{(1) Inputs \& Outputs}: Input refers to the textual content from a benchmark that is provided to the LLM in a single evaluation instance, while the output includes standardized status codes, generated response, error logs, and metadata. 
\textbf{(2) Communication Protocol}: The LLM adapter and client interact via a unified JSON communication format and standardized status codes. The overall communication style follows \texttt{RESTful HTTP} semantics, with behavioral semantics defined in reference to \texttt{RFC 2119} constraints. For example, a 200 status code \texttt{MUST NOT} be retried, while a 429 code \texttt{SHOULD} trigger a reduction in concurrency and a backoff retry. Through this specification, the client can automatically execute retry, rate-limiting, or termination strategies based on the status, avoiding inconsistent behaviors caused by discrepancies in APIs' native status codes. 
\textbf{(3) Model Management}: A Model Card serves as metadata to declare model capability, parameter size, and API endpoints, enabling an automatic discovery mechanism. By inheriting from a base model class, all models share protocol logic, such as logging and unified return formats, ensuring seamless LLM integration.

\subsection{Benchmark Server}

\begin{figure}[!t]
    \centering
    \includegraphics[width=\linewidth]{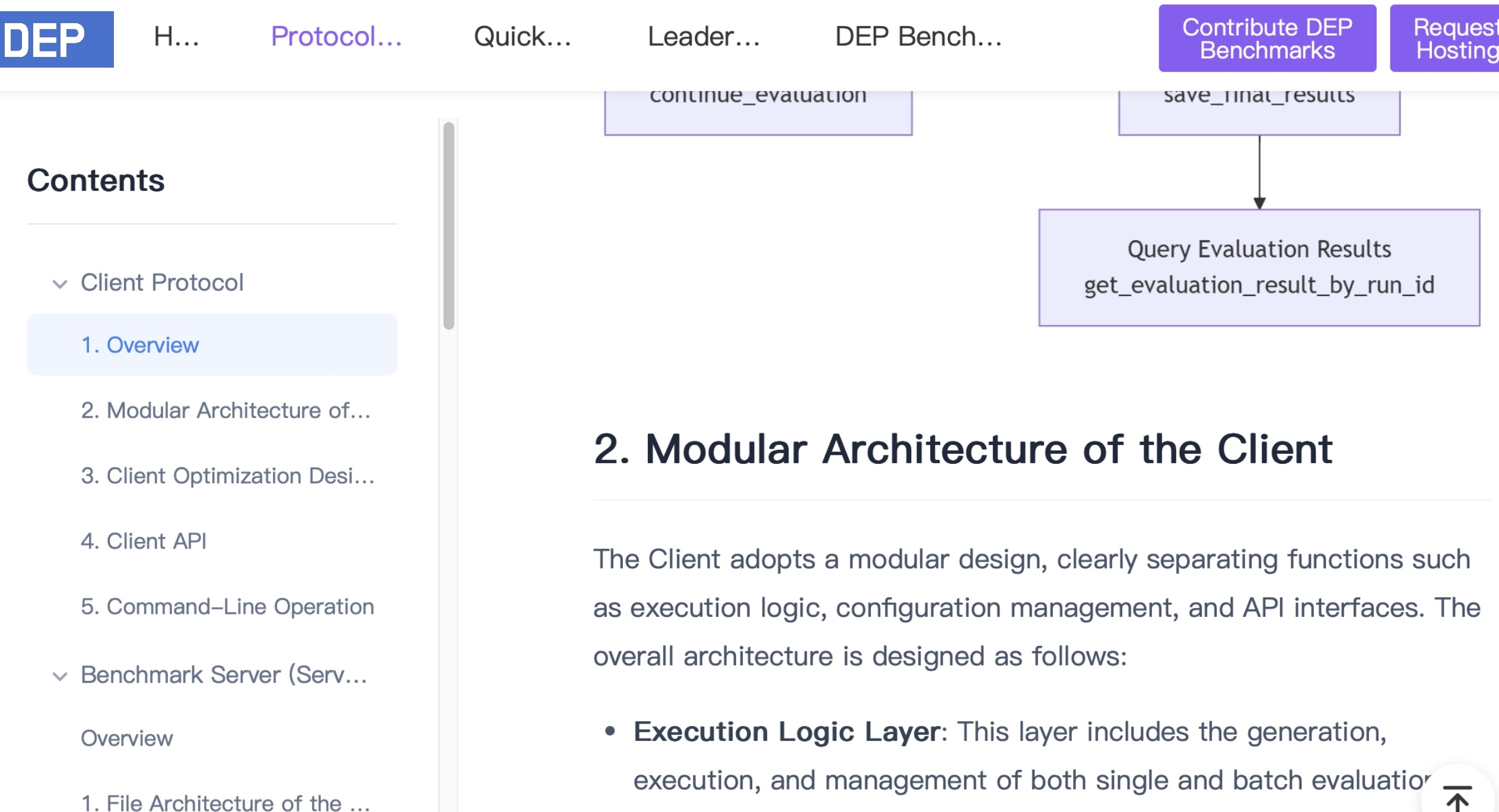}
    \caption{The screenshot of DEP website.}
    \label{Figwebdocs}
\end{figure}

In DEP, each benchmark is packaged as an independent benchmark server that contains a data loading module and an evaluator module, and exposes interface specifications and configurations through a unified Dataset Card metadata file.

DEP allows developers to encapsulate existing benchmarks into DEP without changing the original dataset format. The server connects to raw benchmark files (e.g., Parquet, JSONL, CSV, and custom structures), and standardizes disparate file formats and structures into a unified format for LLM input through its interface. This input follows the data loading module’s preset logic, such as the chain of thought (CoT) prompt set by the author, thus ensuring the reproducibility of the evaluation.

During evaluation, the evaluation module loads the LLM-generated outputs submitted by client and the ground-truth answers, applies predefined post-processing methods to the LLM outputs, and executes the evaluation logic internally. The evaluation interface returns structured results including overall performance, sub-task performance, and optional detailed scores. In addition to common metrics, developers are permitted to use custom evaluation scripts. The system is also compatible with Hugging Face Evaluate, LLM as a Judge or other external evaluation tools. Since all evaluation logic and ground-truth answers are strictly confined to the server side, neither the client nor the LLMs can access it, effectively preventing answer leakage and data contamination at the architectural level.

\subsection{Client}

The client serves as the coordination layer between the LLM adapter and benchmark, responsible for task orchestration and lifecycle management. The task lifecycle states include: \texttt{running}, \texttt{paused}, \texttt{completed}, and \texttt{failed}, all state transitions are uniquely identified by an \texttt{Evaluation ID}. 

The client also supports automatic discovery of benchmarks and models by scanning directories for Dataset Cards and Model Cards to load server and adapter configurations, enabling plug-and-play expansion.

\section{DEP Toolkit}

We develop \textbf{DEP toolkit}, a comprehensive evaluation client based on DEP. The following sections outline the primary functional implementations of DEP toolkit. Usage instructions are available in documentation at online DEP website.\footnote{http://dep.openeval.org.cn/ProtocolDocs}

\paragraph{User Interaction.} The DEP toolkit provides complete evaluation lifecycle management. The toolkit can be installed via:

\begin{tcolorbox}[terminalstyle]
\texttt{\$ pip install llmdep}
\end{tcolorbox}

Users can initiate new evaluation tasks through a command-line interface (CLI). During the evaluation process, users can monitor task status in real-time, including current progress, execution phases, and error logs. If a task is terminated due to external interruptions, the system supports breakpoint resumes based on \texttt{Evaluation ID}, preventing redundant computation and token wastage. The Client also supports unified querying and aggregated display of results, allowing users to view detailed metrics for a single model on a specific benchmark or perform horizontal comparisons across multiple models to generate leaderboards clustered by task dimensions.

\paragraph{System Management.} The DEP toolkit adopts a modular architectural design, decoupling evaluation control logic, configuration management, API layers, log system, and storage. The automatic discovery mechanism enables the system to scan directories and dynamically load compliant benchmarks and LLMs. When a new LLM or benchmark is added, it can be integrated simply by loading it into a designated directory, a plug-and-play approach that enhances scalability and reduces operational overhead. Furthermore, all results and intermediate states are stored in structured formats locally.

\paragraph{Evaluation Controller.} DEP toolkit implements task scheduling and traffic-control mechanisms tailored for large-scale LLM evaluation. It supports both concurrent and batch execution, with configurable parameters to control the execution granularity of generation and evaluation. For API-based inference, DEP toolkit features a built-in congestion control mechanism based on the Token Bucket Algorithm. This is used to smooth request traffic and limit instantaneous bursts, preventing the frequent triggering of API rate limits. Additionally, the system maintains detailed logs of request history, rate-limiting events, and exception types.

\paragraph{Built-in Metrics.} Our toolkit provides a variety of preset metrics for benchmarking (e.g., ACC, EM, F1), which developers can use directly to facilitate DEP server construction.

\section{Evaluation Ecology Strategy}
We further propose the DEP evaluation ecology strategy centered on benchmark lifecycle management and data governance. The strategy addresses two core challenges in LLM evaluation: benchmark fragmentation, test-set contamination and evaluation unsustainability.

\subsection{Collaborative Construction and Sharing}

We continuously adapt benchmarks to DEP and release their implementations to foster reuse. Building on this foundation, we will conduct regular LLM evaluations and publish comprehensive reports to systematically track model development trends in knowledge coverage, reasoning capabilities, and safety alignment.

Furthermore, as a decentralized system, DEP serves as a protocol ecosystem for the LLM community. We encourage benchmark authors to contribute their adaptation implementations. For benchmarks with existing DEP adaptations, users can integrate them into the client and use them with zero code. This low-cost method enables evaluation organizers to seamlessly introduce new benchmarks, significantly shortening the time for incorporating new tasks and datasets.

\subsection{Sustainable Evaluation}
To prevent LLMs exposing to publicly available benchmark answers during pretraining or alignment stages, researchers commonly adopt online evaluation settings, where users submit LLM outputs, and the evaluation organizers score and release the final results. DEP naturally supports this evaluation paradigm. Since both the ground truth and the evaluation logic reside exclusively on the benchmark, the client has no access to the ground truth. This structural separation helps prevent benchmark contamination and answer leakage. In Appendix \ref{sec:appendixremote}, we present a use case demonstrating how private deployment of the server enables such confidential evaluation in practice.

\begin{figure*}[!t]
    \centering
    \includegraphics[width=\linewidth]{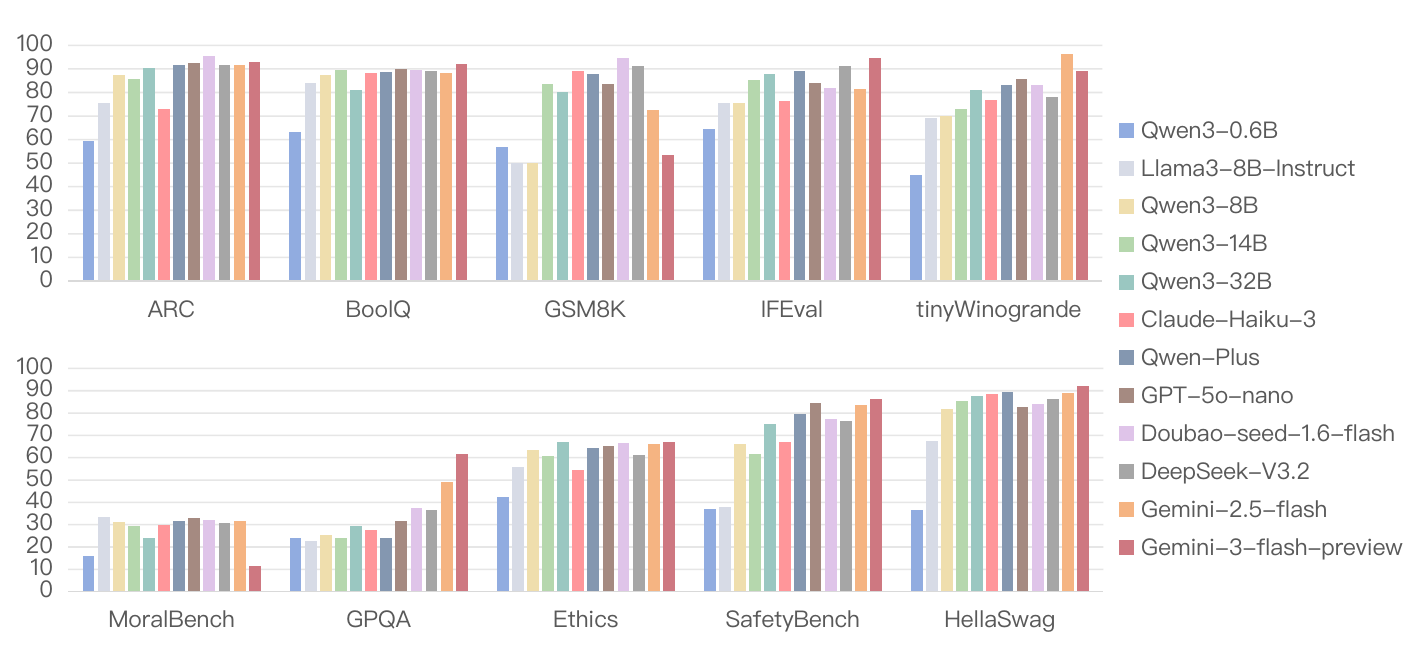}
    \caption{Main results in the LLM evaluation with DEP.}
    \label{fig:mainexp}
\end{figure*}

\section{Experiments}
In January 2026, we conducted an evaluation across 10 benchmarks using 12 open-source and closed-source LLMs via the DEP toolkit to verify the usability of our system. This section presents the primary findings of the assessment. Furthermore, we recruited users to evaluate the utility and ease of use of the DEP. Detailed descriptions of two use cases, covering standard local evaluation and evaluation via a remotely deployed server, can be found in Appendix \ref{sec:appendixcase}.

\subsection{LLM Evaluation}

\paragraph{Setup.} We evaluated 6 open-source LLMs and 6 closed-source LLMs, and the scale of the open-source models ranges from 0.6B to 685B parameters. We utilized computational resources totaling 18 million tokens and 130 GPU hours (NVIDIA A100 80G). For the closed-source LLMs and DeepSeek-V3.2, we utilized API calls. Table \ref{tab,llm} and Table \ref{tab,bench} provide a comprehensive list of all tasks and LLMs used in the evaluation.

\paragraph{Results.} 
 The experimental results are shown in Figure \ref{fig:mainexp}. First, on classic benchmarks such as ARC and BoolQ, the performance gap among most LLMs is relatively small, except for extremely small language models (SLMs) such as Qwen3-0.6B. This suggests that some traditional benchmarks have begun to exhibit saturation, leading to diminished discriminative power in evaluation. In contrast, more complex tasks such as mathematical reasoning show much larger performance gaps and remain core indicators for distinguishing SOTA LLMs.
 
 In addition, the results show a clear scaling effect. Across nearly all tasks, the Qwen3 series exhibits a consistent performance improvement as LLM size increases. From 0.6B to 32B parameters, scores rise almost monotonically across benchmarks, indicating that scale remains a key factor in solving complex problems. However, somewhat unexpectedly, Gemini-3-flash-preview performs relatively poorly on the GSM8K task. Although it is able to follow the CoT format and produce correctly structured outputs, it still generates incorrect answers, suggesting the presence of certain deficiencies in its mathematical reasoning capability.
 
 Finally, although larger-scale LLMs demonstrate stronger knowledge and capabilities, they do not necessarily outperform SLMs in terms of ethical alignment and safety. This suggests that, while continuously improving LLM capabilities, we must also pay close attention to ethical alignment and potential safety risks. These findings demonstrate the effectiveness of DEP-based evaluation.

\subsection{User Study}
We conducted a user study involving 8 participants across 12 benchmarks. The study compared four different methods: (1) Manually implementing evaluation scripts, (2) Adapting to the evaluation framework: OpenEval, (3) Adapting to DEP, and (4) Reusing an existing DEP benchmark. We measured the average time consumption and average lines of code (LoC) required per benchmark. Detailed experimental settings are provided in Appendix \ref{sec:appendixuser}.

As shown in Table \ref{tab,user}, users and benchmark authors can adapt a pluggable DEP server without increasing their workload (compared to manual implementation or evaluation framework adaptation). Once a DEP server is available, users can conveniently mount the benchmark with zero-code effort.

\begin{table}[t]
\centering
\begin{tabular}{lcccc}
\toprule
\textbf{Metric}          & \textbf{MI} & \textbf{A2E} & \textbf{A2D} & \textbf{RD} \\
\midrule
Time Cost (min) & 45                    & 92                               & 69               & 3                                  \\
LoC             & 109                   & 215                              & 162              & 0  \\
\bottomrule
\end{tabular}
\caption{Average Time Cost and lines of code (LoC) for preparing a single benchmark. MI: Manual Implementation; A2E: Adapting to Evaluation Framework; A2D: Adapting to DEP; RD: Reusing an Existing DEP Benchmark.}
\label{tab,user}
\end{table}

\section{Conclusions}
We have presented DEP, which provides a unified, plug-and-play interface while decoupling benchmarks from evaluation clients to support data isolation. Furthermore, we release DEP toolkit, a protocol-compliant evaluation tool that addresses practical needs for large-scale benchmarking, including task management, resource scheduling, fault recovery. By integrating standardized protocol interfaces with systematic tooling capabilities, we construct an extensible, reproducible, and sustainable benchmark evaluation protocol. This
supports for more convenient and standardized LLM evaluation. Our experimental results demonstrate the effectiveness of evaluation using DEP. Compared with other adaptation approaches, it requires no additional workload to complete the adaptation process, and once adapted, it can be conveniently reused.

\clearpage

\section*{Limitations}
The overall effectiveness of DEP depends on the number of protocol-compliant servers available. While we are committed to continuously adapting benchmarks to the DEP, contributions from benchmark authors and developers are crucial, which inevitably introduces an additional integration workload. Currently, DEP is focused on LLM evaluation and cannot yet evaluate agents involving complex external interactions, such as executing software development tasks in Docker environments. This represents a future direction for DEP's expansion.


\bibliography{custom}

\begin{figure*}[!t]
    \centering
    \includegraphics[width=\linewidth]{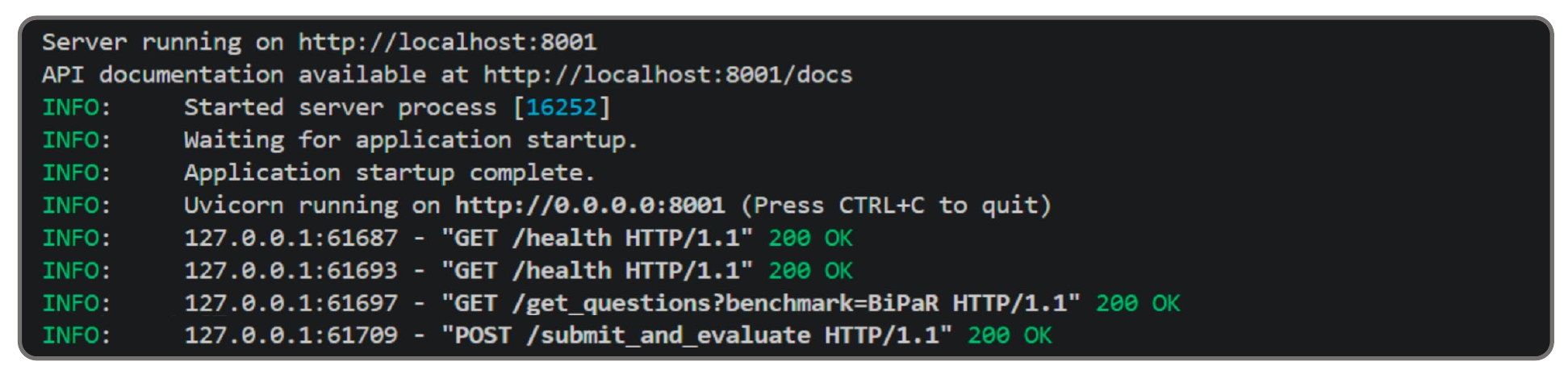}
    \caption{Server-side interface for remote evaluation.}
    \label{remoteserver}
\end{figure*}

\begin{figure*}[!t]
    \centering
    \includegraphics[width=\linewidth]{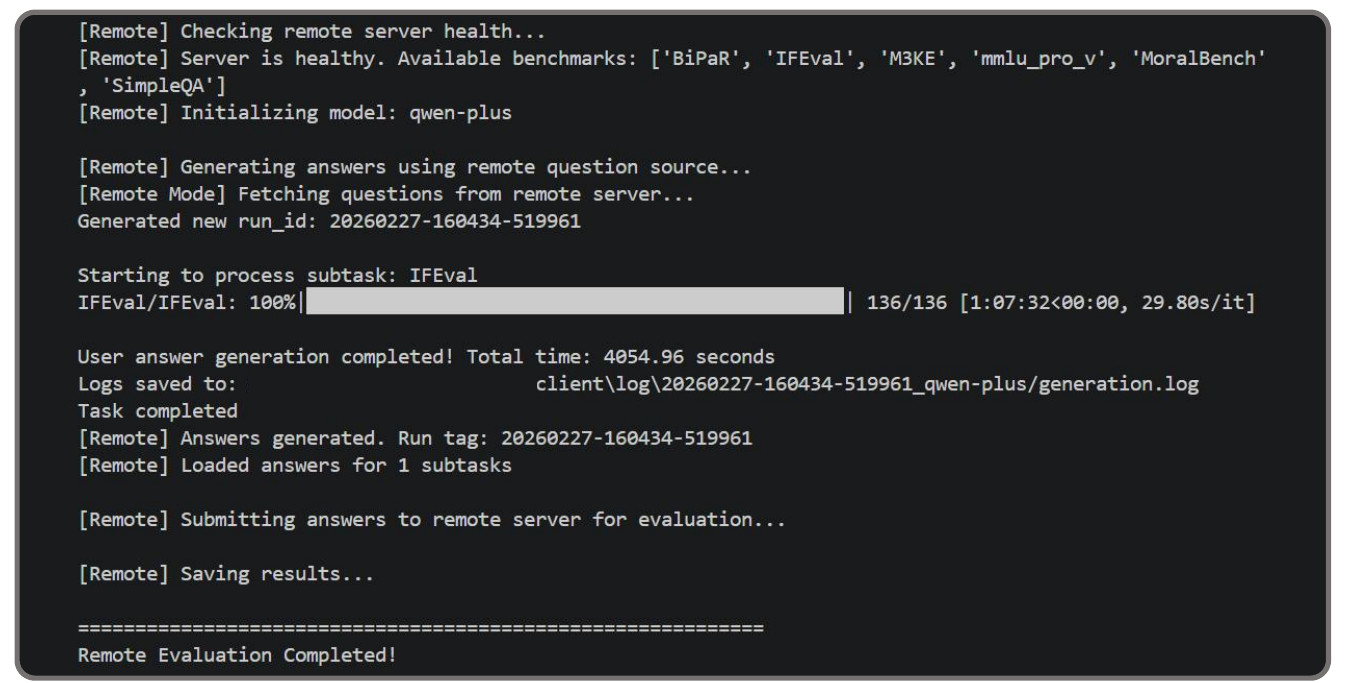}
    \caption{Client-side connection to the remote benchmark server.}
    \label{remoteclient}
\end{figure*}

\appendix

\section{Case Study}
\label{sec:appendixcase}
\subsection{Regular Evaluation}
\begin{figure*}[!t]
    \centering
    \includegraphics[width=\linewidth]{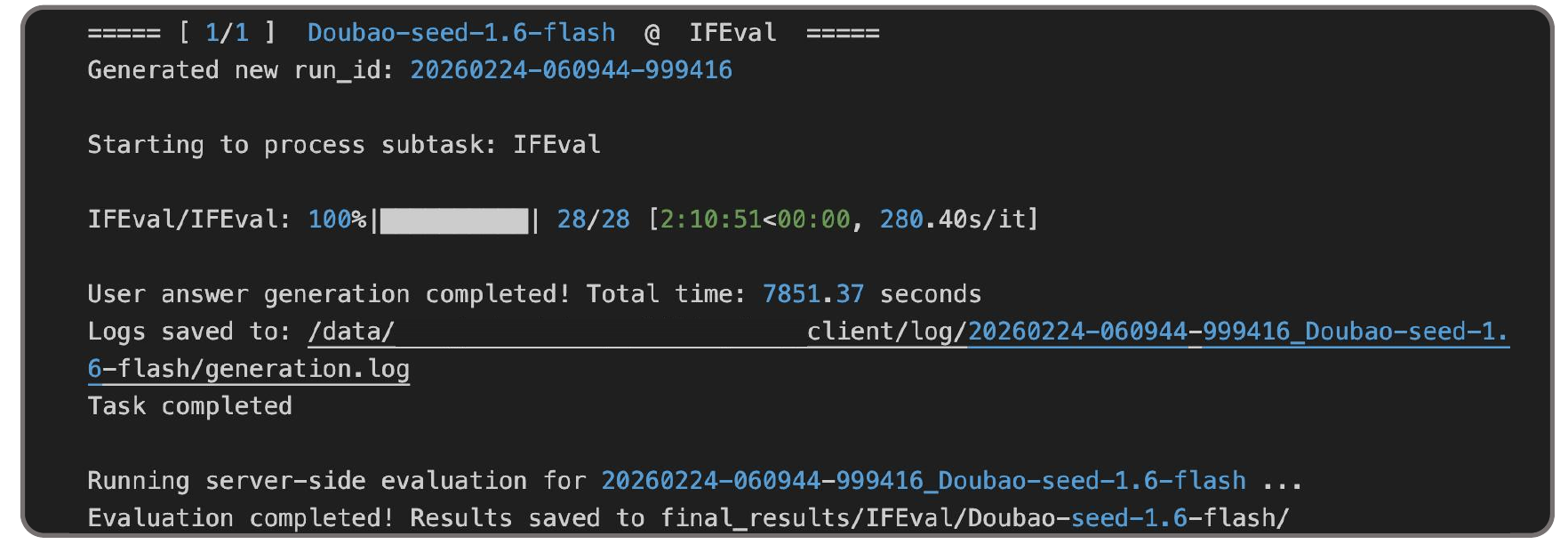}
    \caption{Use case of a regular evaluation with DEP toolkit.}
    \label{case3}
\end{figure*}

Figure \ref{case3} illustrates a typical evaluation using the DEP toolkit. We conducted an assessment on the IFEval dataset using the Doubao-seed-1.6-flash model. The screenshot depicts the execution logs, covering the entire workflow from dataset loading and evaluation execution to result submission and final scoring.

\begin{table*}[t]
\centering
\begin{tabular}{ccccc}
\toprule
\textbf{ID} & \textbf{Type} & \textbf{Model Name}                & \textbf{\#Param.}  & \textbf{Access}  \\
\midrule
1  & Closed-source       & Claude-Haiku-3 \citep{claude3}& -         & API     \\
2  &  Closed-source       & GPT-5o-nano \citep{gpt5}              & -         & API     \\
3  &  Closed-source       & Gemini-2.5-flash \citep{comanici2025gemini}         & -         & API     \\
4  &  Closed-source       & Gemini-3-flash-preview \citep{gemini3}   & -         & API     \\
5  &  Closed-source       & Doubao-seed-1.6-flash \citep{seed}    & -         & API     \\
6  &  Closed-source       & Qwen-Plus \citep{bai2023qwen}                & -         & API     \\
7  &  Open-source       & DeepSeek-V3.2 \citep{deepseek3-2}            & 37/785B   & API     \\
8  &  Open-source       & Qwen3-0.6B \citep{yang2025qwen3}               & 0.6B      & Weights \\
9 &  Open-source       & Qwen3-8B \citep{yang2025qwen3}                 & 8B        & Weights \\
10 &  Open-source       & Qwen3-14B \citep{yang2025qwen3}                & 14B       & Weights \\
11 &  Open-source       & Qwen3-32B \citep{yang2025qwen3}                & 32B       & Weights \\
12 &  Open-source       & Llama3-8B-Instruct \citep{dubey2024llama}       & 8B        & Weights\\
\bottomrule
\end{tabular}
\caption{LLMs used in Section 6.1.}
\label{tab,llm}
\end{table*} 

\subsection{Remote Deployment}
\label{sec:appendixremote}
If benchmark developers wish to conduct evaluations in a private manner (i.e., without disclosing the Ground Truth), they need to deploy the server and open the corresponding ports. We simulated the deployment of six benchmark servers for testing, including BiPaR, IFEval, M3KE, MMLU\_Pro, MoralBench, and SimpleQA.

As shown in Figure \ref{remoteserver}, this is the server side of the BiPaR dataset, which accepts evaluation task submissions from the client. Figure \ref{remoteclient} illustrates the client side, displaying six deployed and available benchmark servers and the execution and submission of an evaluation.

\section{User Study Details}
\label{sec:appendixuser}

We recruited 8 computer science professionals with undergraduate degrees and backgrounds in LLM research, compensating them at a rate of \$9 per hour. We compared four distinct methods: manual implementation, adapting to the OpenEval platform, adapting to a DEP Server, and utilizing an existing DEP Server.

\textbf{Manual Implementation.} 1. Research the benchmark paper and data files. 2. Write a serial evaluation script including basic test set traversal, result storage, and evaluation logic. 3. Verify and test the script. 4. Execute the evaluation.

\textbf{Adapting to OpenEval.} 1. Research the benchmark paper and data files. 2. Reorganize and adapt the benchmark to meet the platform's data format and evaluation methodology requirements. 3. Verify and test the script. 4. Execute the evaluation via the platform.

\textbf{Adapting to DEP.} 1. Research the benchmark paper and data files. 2. Adapt the benchmark based on a DEP template script following the protocol documentation. 3. Verify and test the script. 4. Execute the evaluation using the DEP toolkit.

\textbf{Using an Existing Benchmark Server.} 1. Load the target benchmark server into the DEP toolkit. 2. Execute the evaluation using the DEP toolkit.

For the first three methods, we recorded the average time and lines of code required for steps 2 and 3. For the fourth method, we recorded these metrics for steps 1 and 2. All code underwent cross-verification to ensure correctness.

\begin{table}[t]
\centering
\begin{tabular}{cc}
\toprule
\textbf{ID} & \textbf{Benchmark}     \\
\midrule
1  & GSM8K \citep{cobbe2021trainingverifierssolvemath}         \\
2  & GPQA \citep{rein2024gpqa}          \\
3  & IFEval \citep{cobbe2021trainingverifierssolvemath}        \\
4  & MoralBench \citep{ji2025moralbench}    \\
5  & ARC \citep{clark2018think}           \\
6  & Ethics \citep{hendrycks2020aligning}        \\
7  & BoolQ \citep{clark2019boolq}         \\
8  & HellaSwag \citep{zellers2019hellaswag}     \\
9  & tinyWinogrande \citep{polo2024tinybenchmarks}\\
10 & SafetyBench \citep{zhang-etal-2024-safetybench}\\  
\bottomrule
\end{tabular}
\caption{Benchmarks evaluated in Section 6.1.}
\label{tab,bench}
\end{table}

\end{document}